\documentclass{article}

\usepackage{microtype}
\usepackage{graphicx}
\usepackage{subfigure}
\usepackage{multirow}
\usepackage{booktabs} 
\usepackage{authblk}

\usepackage{hyperref}

\usepackage[accepted]{icml2024}

\usepackage{amsmath}
\usepackage{amssymb}
\usepackage{mathtools}
\usepackage{amsthm}
\usepackage{stfloats} 

\usepackage[capitalize,noabbrev]{cleveref}

\usepackage{url}

\theoremstyle{plain}

\theoremstyle{definition}

\theoremstyle{remark}

\usepackage[textsize=tiny]{todonotes}

\icmltitlerunning{Distilling Transformers into Long Convolution Models}

\begin{document}

\twocolumn[
\icmltitle{Scavenging Hyena: Distilling Transformers into Long Convolution Models}



\icmlsetsymbol{equal}{*}

\begin{icmlauthorlist}
\icmlauthor{Tokiniaina Raharison Ralambomihanta}{mcgillaff}
\icmlauthor{Shahrad Mohammadzadeh}{mcgillaff}
\icmlauthor{Mohammad Sami Nur Islam}{mcgillaff}
\icmlauthor{Wassim Jabbour}{mcgillaff}
\icmlauthor{Laurence Liang}{mcgillaff}
\end{icmlauthorlist}

\icmlaffiliation{mcgillaff}{McGill University, Montreal, Canada}

\icmlcorrespondingauthor{Tokiniaina Raharison Ralambomihanta}{tokiniaina.raharisonralambomihan@mail.mcgill.ca}


\icmlkeywords{Machine Learning, ICML}
\vskip 0.3in
]



\printAffiliationsAndNotice{}  

\begin{abstract}

The rapid evolution of Large Language Models (LLMs), epitomized by architectures like GPT-4, has reshaped the landscape of natural language processing. This paper introduces a pioneering approach to address the efficiency concerns associated with LLM pre-training, proposing the use of knowledge distillation for cross-architecture transfer. Leveraging insights from the efficient Hyena mechanism, our method replaces attention heads in transformer models by Hyena, offering a cost-effective alternative to traditional pre-training while confronting the challenge of processing long contextual information, inherent in quadratic attention mechanisms. Unlike conventional compression-focused methods, our technique not only enhances inference speed but also surpasses pre-training in terms of both accuracy and efficiency. In the era of evolving LLMs, our work contributes to the pursuit of sustainable AI solutions, striking a balance between computational power and environmental impact.
    
\end{abstract}

\section{Introduction}


In recent years, the field of natural language processing (NLP) has been revolutionized by the advent of Large Language Models (LLMs), with the transformer architecture, introduced in 2017 by \citeauthor{Attention_is_all_you_need}, marking a significant turning point in the literature. Despite the lack of a universally accepted definition for LLMs, they can be broadly conceptualized as robust machine learning models capable of executing a multitude of natural language processing tasks simultaneously. As delineated by \citeauthor{yang2023harnessing} in 2023, these tasks encompass:

\begin{enumerate}
    \item Natural language understanding
    \item Natural language generation
    \item Knowledge-intensive tasks
    \item Reasoning ability
\end{enumerate}


Indeed, the landscape of Large Language Models (LLMs) has seen a proliferation of diverse architectural strategies. These encompass models that leverage both encoders and decoders, models that solely employ encoders such as BERT, and models that are exclusively decoder-based like GPT-4. It has been observed that decoder-only models, exemplified by GPT-4, demonstrate superior performance, especially in tasks pertaining to natural language generation, when juxtaposed with their encoder-based counterparts. This suggests a potential trend towards decoder-only models in the pursuit of enhanced performance, especially when it comes to natural language generation tasks.

In the preceding year, OpenAI introduced the GPT-4 Turbo model, a significant advancement over its predecessors in terms of performance \cite{OpenAI_2023}. However, the GPT-4 model, with its approximately 1.7 trillion parameters, has sparked concerns about the substantial energy resources necessitated for its pre-training. This underscores the importance of developing sustainable AI solutions that balance computational power and environmental impact.

Our research explores the concept of distillation as a proficient methodology for training Large Language Models (LLMs) with new architectures. This approach aims to mitigate the substantial electricity consumption and financial expenditure associated with the pre-training of new architectures, especially when the knowledge of other pre-trained LLMs can be utilized.

In particular, our work investigates distilling the knowledge of an LLM that uses traditional, quadratic multi-headed attention into an equivalent model that uses sub-quadratic Hyena operators instead \cite{poli2023hyena}. It then proceeds to compare the results of the distillation to training that latter model from scratch.

Our work also addresses the need for models to efficiently process long context lengths, as a longer context length correlates to larger model memory and more complex model reasoning \cite{ding2023longnet}. The quadratic nature of attention mechanisms poses a fundamental challenge in traditional models, limiting their ability to effectively incorporate long contextual information. Recognizing the inherent advantages of utilizing longer context in understanding and generating meaningful sequences, it becomes crucial to overcome the quadratic scaling issue.

In traditional distillation approaches, the primary focus is on enhancing inference speed through the compression of existing models into more compact versions of the same architecture. However, a notable drawback of this method is its tendency to diminish the language modeling abilities of the model. Moreover, the approach does not address the quadratic scaling issue in length, as maintaining the same architecture fails to resolve the long context problem. \textbf{Our research addresses these limitations by proposing a novel approach using knowledge distillation methods to efficiently transfer knowledge from existing transformers into long convolution models, creating a model that exhibits improved scaling concerning context length as well as reduced training costs when compared with the standard pre-training approach.} The following points describe the main approaches towards achieving the desired efficiency:

\begin{itemize}
    \item Knowledge Distillation for Cross-Architecture Transfer: Our research pioneers a novel approach by employing knowledge distillation techniques not only for model compression but also for transferring knowledge from existing transformers to long convolution models. 
    \item Knowledge Distillation Surpassing Pre-training Efficiency:
Our research establishes a superior distillation paradigm, outperforming traditional pre-training both in terms of accuracy and efficiency.
\end{itemize}




\textbf{}

\section{Background}

\subsection{Self Attention Mechanism}
In transformers, for a length-\(L\) sequence \(u \in \mathbb{R}^{L \times D}\), the scaled self-attention mechanism involves three learnable linear projections \(M_q, M_k, M_v \in \mathbb{R}^{D \times D}\). These projections are applied to the input sequence \(u\) to compute Query (\(Q\)), Key (\(K\)), and Value (\(V\)) matrices:

\[Q = u \cdot M_q, \ K = u \cdot M_k, \ V = u \cdot M_v.\]

The attention operation is defined as follows:

\[A(u) = softmax \left( \frac{QK^T}{\sqrt{D}} \right),\]

where SoftMax is applied row-wise. The output of self-attention \(y\) is obtained by multiplying the attention weights \(A(u)\) with the Value matrix \(V\):

\[y = \text{SelfAttention}(u) = A(u) \cdot V.\]

This mechanism enables the model to capture dependencies among elements in the input sequence, assigning varying importance to different elements during computations. By learning to attend to relevant parts of the sequence, self-attention enhances the model's ability to process sequential data efficiently.

\subsection{Subquadratic Attention Replacements}

The challenge with standard attention \cite{Attention_is_all_you_need} lies in its quadratic scaling with input length $N$, prompting the exploration of subquadratic alternatives. Notable examples include the Attention Free Transformer \cite{zhaiAttentionFreeTransformer2021a} and linear attention \cite{katharopoulosTransformersAreRNNs2020a}, where the time complexity is reduced while maintaining the overall integrity of the transformer architecture.

Another alternative to attention is the use of state space models where we capture the dynamics of the system through difference equations. These models use linear mappings from an input signal to an output signal where the output signal $y[n]$ is a function of the input signal $u[n]$ and a state variable $x[n]$:

\[
    \begin{aligned}
        x[n+1] &= A x[n] + B u[n]\\
        y[n] &= C x[n] + D u[n]
    \end{aligned}
\]

The state space representation provides a direct means of computing the output through the recurrence relationship. Enforcing linearity and time variance allows us to equivalently compute the output $y[n]$ through a convolution with the system's impulse response $h[n]$:

\[
    y[n] = u[n] * h[n] = u[n] * (CA^nB + D\delta[n])
\]

where $*$ denotes the convolution operation, and $\delta$ the Kronecker delta function. This convolution view lets us efficiently compute the output in $O(N (\log{N})^2)$ through the fast Fourier transform algorithm \cite{brighamFastFourierTransform1967}. Consequently, one can opt to parameterize $A,B,C,D$ directly as structured matrices, as demonstrated in \cite{fuHungryHungryHippos2022}. Alternatively, Hyena \cite{poli2023hyena} introduces a novel approach with the parametrization of an implicit long convolution, which can then be distilled into a state space representation for constant time inference \cite{massaroliLaughingHyenaDistillery2023}.

\subsection{Distillation}

Knowledge distillation in neural networks \cite{Hinton2015DistillingTK} involves transferring information from a larger, more complex model to a smaller one while minimizing information loss. This method extends to both compressing a single larger model and consolidating insights from multiple models (ensemble) into a singular one.

Distillation, a knowledge transfer method in neural networks, leverages temperature-adjusted softmax probabilities. Initially, the cumbersome model generates soft targets by applying a higher temperature in its softmax, aiding the training of a smaller distilled model. Besides mimicking soft targets, optimizing the distilled model with correct labels further enhances learning.

The training involves a weighted average of two objective functions: the first part is the Kullback–Leibler divergence with the soft targets (at higher temperature). The second part is the cross entropy loss with correct labels (at temperature 1). 

This methodology allows the distilled model to effectively learn from both the nuanced information present in the soft targets generated by the larger model and the precise ground truth labels, resulting in a more compact yet knowledgeable model.

One notable example of distillation in LLMs is the DistilBERT model: DistilBERT is 40\% smaller than its parent model BERT, 60\% faster than its parent model, and yet retains 97\% of BERT's language capabilities. \cite{sanh2020distilbert} 


\subsection{Progressive Knowledge Transfer.} 
\begin{figure*}[h]
    \centering
    \includegraphics[width=0.8\linewidth]{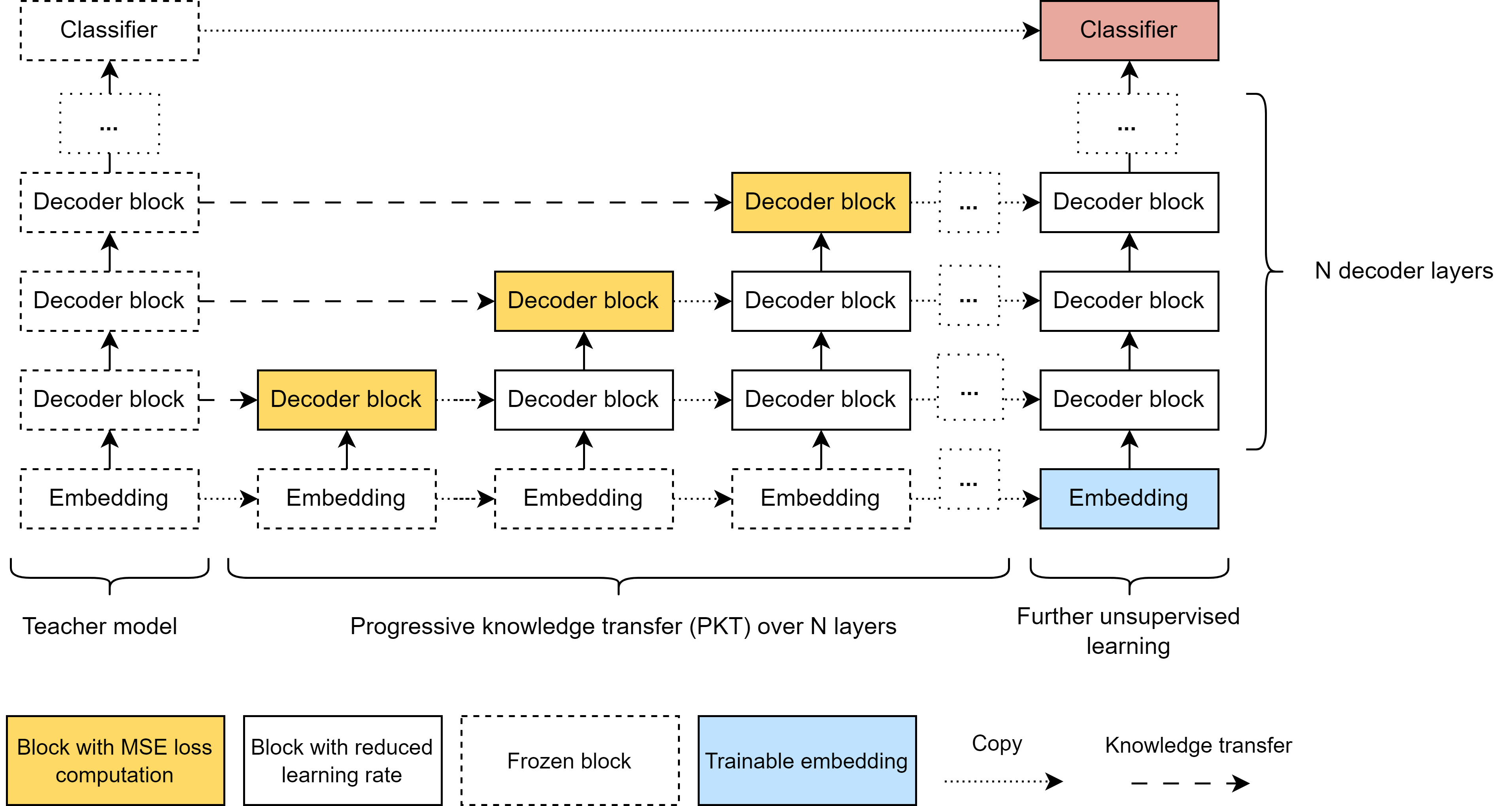}
    \caption{Progressive knowledge transfer on a Pythia model on its decoder layers. Adapted from \cite{sun2020mobilebert}.}
    \label{fig:knowledge-transfer}
\end{figure*}

When distillation is implemented on large models, there is a risk that knowledge transfer is not optimally passed on from the teacher model to the student model due to differences between the architectures of the teacher and student models. One approach to maximize knowledge transfer is progressive knowledge transfer: the student model is first trained only on the inputs and outputs of the first encoder block, and the student model then subsequently trains the output of the next encoder block while freezing the previous trained blocks.  \cite{sun2020mobilebert} In our case, encoder blocks are replaced by decoders as the architecture is autoregressive. (Fig. \ref{fig:knowledge-transfer})

\section{Methods}

\subsection{Hyena Operator}

Hyena \cite{poli2023hyena} proposes the use of implicit long convolutions as a subquadratic replacement for the attention operator. Instead of parametrizing the state space coefficients as in other state space models such as H3 \cite{fuHungryHungryHippos2022}, it chooses to directly parametrize filters $h: \mathbb{N} \to \mathbb{R}^{d} $ --- equivalent to an LTI system's impulse response. The filter is obtained by first applying a positional embedding $P_e: \mathbb{N} \to \mathbb{R}^{d_f}$ --- where $d_f$ is the embedding dimension --- to the time indices. We then apply a feed forward neural network $\text{FFN}: \mathbb{R}^{d_f} \to \mathbb{R}^{d_m}$ --- where $d_m$ is the model's dimension --- and multiply by a windowing function to obtain the filter.

\[
    h[n] \coloneq \text{Window}( \text{FFN}(P_e[n]))
\]

The hyena operator $H: \mathbb{R}^{d_m} \to \mathbb{R}^{d_m}$ uses one such filter $h$ to aggregate context over a long context window and adds non-linearity through a multiplicative gating mechanism. The first step is to obtain three projections $q,k,v$ through the projection operation $P(x,\theta)$ with parameters $\theta$. The projection operations consist of a linear projection $W_\theta$ followed by a short depth-wise convolution with a short filter $k_\theta$ for local information exchange.
We then use an element wise multiplication followed by a convolution and a second element wise multiplication to compute the output of the hyena operator:

\[
    \begin{aligned}
        P_\theta(x) &\coloneq k_\theta * (x \cdot W_{\theta})\\ 
        H(x) &\coloneq P(x;\theta_q) \odot (h * (P(x;\theta_k) \odot P(x;\theta_v)))
    \end{aligned}
\]

where $*$ is the convolution operation and $\odot$ is the element-wise multiplication. Note that the operator can be further generalized by using different numbers of projections \cite{poli2023hyena}.

\subsection{Model}

\begin{figure*}[h]
  \centering
  \includegraphics[width=0.6\linewidth]{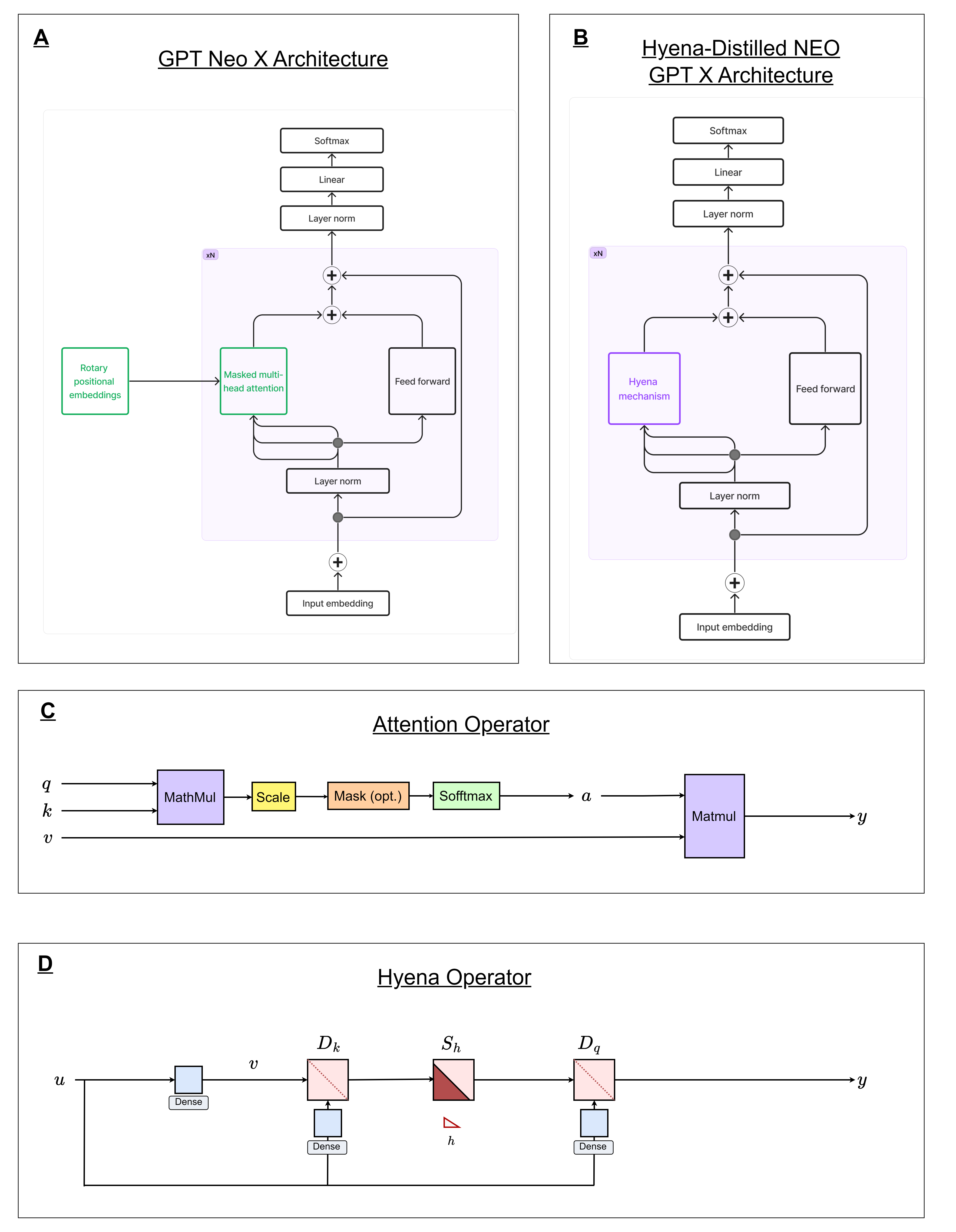}
  \caption{\textbf{(A)} GPT NEO X Layer Architecture: 6 layers of stacked Attention and MLPs in the 70M GPT NEO X. \textbf{(B)} Hyena-Distilled NEO GPT X Layer Architecture: Replacement of attention heads by the Hyena operator for the distillation task. \textbf{(C)} A visual representation of the attention operator, adapted from \cite{Attention_is_all_you_need}. \textbf{(D)} A visual representation of the Hyena operator, adapted from \cite{poli2023hyena}.}
  \label{fig:model-comparisons}
\end{figure*}

In terms of the model used to conduct our experiments, we opted for the 70M parameter version of GPT-NeoX \cite{black2022gptneox20b}, which is a decoder-only transformer model whose architecture closely matches that of GPT-3, except for a few key differences:
\begin{itemize}
    \item {
        The positional embeddings traditionally found in GPT models are swapped for rotary positional embeddings (RoPE), which encode the positional information of tokens using a rotation matrix.
    }
    \item {
        The attention and feed-forward layers that are usually found in series in traditional GPT models are instead computed in parallel for efficiency purposes.
    }
    \item {
        All feed-forward layers are dense, contrary to the alternance of dense and sparse layers in GPT-3.
    }
\end{itemize}
It is useful to note that the GPT-NeoX architecture closely matches that of GPT-J. Figure \ref{fig:model-comparisons} displays a detailed diagram of the architecture of the model. For the purposes of this paper, the goal was to replace the attention mechanism with a Hyena mechanism, as displayed in Figure \ref{fig:model-comparisons}. It is, however, important to note that the Hyena version of the model does not incorporate rotary positional embeddings due to the fact that the Hyena operator already retains positional information about its input tokens. Finally, we used the Pythia \cite{pythia-paper} implementation of the aforementioned model, trained on the open-sourced Pile \cite{gao2020pile} dataset.



\subsection{Distillation Procedure}

We opt for Progressive Knowledge Transfer \cite{sun2020mobilebert} to progressively train the student model $S(\cdot;\Theta_s)$. For each layer, we first do inference on the teacher model $M(\cdot;\Theta_t)$ over a token dataset $X$ to obtain a distillation dataset $D = \{(x,y_m^i)| x \in X\}$ where $x$ is sequence of token indices and $y^i$ is the teacher model's output at layer $i$. Subsequently, we minimize the mean squared error loss with $y^i_s$—the student model's output at layer $i$ one layer at a time. For the last layer, we can additionally fine tune the model by doing unsupervised training on textual data.
\[
    \begin{aligned}
        \mathcal{L}^i(M(\cdot;\Theta_m),S(\cdot;\Theta_s)) &= \mathbb{E}_{(x,y^i) \sim D} [MSE(y_m^i, y_s^i)]\\
    \end{aligned}
\]

\subsection{Training Dataset and Procedure}

We use OpenWebText \cite{Gokaslan2019OpenWeb} for all language modeling experiments. A tokenized pre-training dataset was obtained by randomly sampling 2M examples from OpenWebText with each pre-training example having a context length of 1024. The dataset was separated into a training set and a validation set with $0.1\%$ being reserved for validation. For distillation experiments, the same 40M tokens were sampled from the training set to obtain the distillation datasets used to train each layer.

All experiments use the same 6-layer GPTNeoX style architecture with the same dimensions as in the 70M teacher model. We first pre-train the model from scratch on 1B tokens based on the hyperparameters for Pythia \cite{pythia-paper} and Hyena models \cite{poli2023hyena}. We define pre-training as the process of doing unsupervised learning on textual data starting with a randomly initialized model. As well, we define unsupervised-tuning (CE-tinetune) as the process of doing unsupervised learning on textual data starting with a model checkpoint.
In our pre-training phase, we implement a linear warm-up spanning 300 training steps, followed by a learning rate decrease using cosine decay over 2000 iterations. This decay continues until we reach $10\%$ of the maximum learning rate, at which point the learning rate remains constant. Similarly, in the distillation process, we incorporate a linear warm-up over $2.5\%$ of the total training steps, followed by a decay over the entire set of steps until we hit $10\%$ of the maximum learning rate. We try only doing distillation (MSE) as well as fine-tuning (CE-tinetune) .All experimment are designed to run in 5 hours on a RTX 3090.

\section{Language Modeling Results}

\subsection{Perplexity Scores}

For OpenWebText, the validation set obtained in the same way as the pre-training dataset was used to compute perplexity for all models. The same procedure was used on the test split of WikiText \cite{merity2016pointer}. The perplexity scores for both WikiText and OpenWebText were obtained over a context length of 1024 tokens.

\begin{table}[h]
\caption{Perplexity scores of Pythia 70M teacher model, pre-trained Hyena model, Hyena student model distilled with MSE loss, and Hyena student model finetuned after distillation from top to bottom respectively.}
\label{perplexities}
\vskip 0.15in
\begin{center}
\begin{small}
\begin{sc}
\begin{tabular}{l c c}
\toprule
\textbf{Model} & \textbf{Wikitext} & \textbf{OpenWebText}  \\
\midrule
Pythia-70m (teacher)   &  51.4  & 35.3 \\
\hline
Pre-trained & 230 & 64.9 \\
MSE & 155.8 & 63.5 \\
CE fine-tune &  \textbf{121.2} & \textbf{49.6}\\
\bottomrule
\end{tabular}
\end{sc}
\end{small}
\end{center}
\vskip -0.1in
\end{table}

\subsection{Language Evaluation}

\begin{table*}[h]
    \caption{Evaluation of Model Performance. Joint knowledge transfer is abbreviated as JKT. All results were measured using the Language Model Evaluation Harness \cite{gao2021framework} with 32-bit floating point precision; the first value is the accuracy, followed by the standard deviation. }
    \label{table:comparison-3-models}
    \vskip 0.15in
    \begin{center}
    \begin{small}
    \begin{sc}
    \begin{tabular}{lcccc}
    \toprule
    \textbf{Task} & \textbf{Metric} & \textbf{GPT Hyena} & \textbf{Pythia 70M Teacher} & \textbf{Pythia 70M JKT Student} \\
    \midrule
    Arc Challenge   & Acc & $0.1775\pm0.0112$ & $0.1749\pm0.0111$ & \boldmath{$0.1792\pm0.0112$} \\
    Arc Easy        & Acc & \boldmath{$0.3998\pm0.0101$} & $0.3754\pm0.0099$ & $0.3270\pm0.0096$\\
    Logiqa          & Acc & $0.1966\pm0.0156$ & \boldmath{$0.2104\pm0.0160$} & $0.1982\pm0.0156$\\
    Piqa            & Acc & $0.5832\pm0.0115$ & \boldmath{$0.5985\pm0.0114$} & $0.5408\pm0.0116$\\
    Sciq            & Acc & $0.5910\pm0.0156$ & \boldmath{$0.6400\pm0.0152$} & $0.3570\pm0.0152$\\
    Winogrande      & Acc & $0.5004\pm0.0141$ & \boldmath{$0.5296\pm0.0140$} & $0.4886\pm0.0140$\\
    Wsc             & Acc & $0.3750\pm0.0477$ & $0.3654\pm0.0474$& \boldmath{$0.5865\pm0.0485$}\\
\bottomrule
\end{tabular}
\end{sc}
\end{small}
\end{center}
\vskip -0.1in
\end{table*}

We applied a series of natural language tasks on three models of interest: (1) a GPT model that used Hyena as a drop-in replacement for attention, (2) a Pythia 70M teacher model that used attention, and (3) a Pythia 70M student model that used Hyena and was distilled via using joint knowledge transfer (JKT).

We used the Language Model Evaluation Harness (lm\_eval) \cite{gao2021framework} to benchmark these three models on multiple different natural language tasks. (Table \ref{table:comparison-3-models}) We used 32-bit floating point precision on all tests to ensure reproducibility and to minimize the effect of machine error due to low precision. 



\section{Discussion}

\subsection{Analysis}

As seen in table \ref{perplexities}, our experimental results demonstrate the advantage of progressive knowledge transfer over traditional pre-training approaches in terms of model performance achieved within a comparable GPU-hour budget. Importantly, without any additional unsupervised learning, our method yields superior performance, indicating the efficiency of our progressive knowledge transfer strategy.

Furthermore, our findings reveal the potential for distillation as an initialization step before unsupervised learning. This approach offers increased performance at the same training cost as conventional pre-training as well as pure knowledge transfer. This suggests that our knowledge distillation approach not only offers improved initial performance but  also allows for additional optimization without incurring additional training expenses.

A closer examination of our results underscores the significant impact of knowledge distillation on model generalization. Indeed, the increased improvements on the WikiText perplexity scores with distillation emphasize the effectiveness of our approach in enhancing the model's capacity to extrapolate on unseen data with the teacher model's knowledge. This contributes valuable insights into the broader applicability and robustness of knowledge distillation in machine learning scenarios, particularly when compared to conventional pre-training strategies.

Table \ref{table:comparison-3-models} suggests that pre-training a GPT model with Hyena generally yields similar yet slightly lower accuracy than a Pythia 70M model that uses Hyena. These results suggest that LLMs that use Hyena are generally able to perform as well as attention-based LLM models, Hyena-based models typically have a slightly lower measured performance. We observe that a student Pythia 70M JKT model generally has a slightly inferior performance compared to a pre-trained attention-based Pythia 70M model, though model performance is generally within a similar range, except for Sciq where the student model's accuracy is noticeably lower than GPT Hyena and the teacher model. However, for the Arc Challenge and Wsc tasks, the Pythia 70M student model slightly outperforms and noticeably outperforms the other two models.

Thus our results suggest that joint knowledge transfer on a student Hyena model generally conserves the language capabilities of its teacher model, and that the student Hyena model can outperform its teacher model in some cases. Because Hyena is more computationally efficient than attention when compared directly, and because joint knowledge transfer may be more computationally efficient than traditional pre-training, our results show encouraging signs that joint knowledge transfer on a Hyena student model offers a computationally efficient alternative to pre-training attention-based large language models. 

\subsection{Limitations}

\textbf{Model Size:} Due to time constraints and limited access to, scaling our approach to larger models was impossible.  Consequently, the generalizability of our approach to deeper or wider models remains unclear. Therefore, further experimentation with larger models remains to be done for assessing the practicality of our method. 

\textbf{Training Time:} Similarly to the above limitation, training times for obtained reported results were limited to 5h. Therefore, we could not determine whether there exists an optimal duration of distillation before normal pre-training becomes advantageous.

\textbf{Benchmarking:} We noticed that using different floating point precision values for the lm\_eval tests would give different results. Thus, we opted to use 32-bit floating point precision, though it is difficult for us to directly quantify how much machine error is present. For the Lambada OpenAI task, some of our models reported a very high perplexity score and a very low accuracy score; we decided to exclude these results from our main results, as further investigation is needed to determine the root cause behind these outlier results. 

\section{Future Work}

In future investigations, we aim to explore the compressibility of the teacher model into a more compact state space model, beyond the current literature's focus on reducing dimensionality and depth. This involves an inquiry into the adaptability of attention mechanisms during compression. Further, we plan to evaluate various distillation approaches, analyzing how performance differences scale with distillation time and the percentage of unsupervised learning. To address the limitations related to model size and training time, future works will involve assessing the proposed approach on larger language models. Additionally, we aspire to evaluate distillation on different sub-quadratic attention replacements, paving the way for a more comprehensive understanding of the applicability and scalability of our knowledge distillation methodology.

\section{Conclusion}
We evaluated the effectiveness of using joint knowledge transfer with Hyena operators (as a drop-in replacement for attention) to improve the computational efficiency of LLMs during training. As a result, we defined a Pythia 70M model with attention as a teacher model, and performed distillation on a Pythia 70M student model by replacing attention with the Hyena operator. By evaluating model perplexity scores on the OpenWebText and WikiText datasets, we observed that a Pythia 70M Hyena model that underwent progressive knowledge transfer performed better than a Pythia 70M Hyena model that was pre-trained. In addition, we observed that fine-tuning Pythia 70M after progressive knowledge transfer noticeably decreases the perplexity score, thus further improving model performance. In terms of natural language tasks, a student Hyena model generally had slighly lower accuracy than its teacher model, though in two instances the student Hyena model was able to outperform its teahcer model. These initial results show encouraging signs that joint knowledge transfer on Hyena student models is capable of conserving a large proportion of a teacher model's langauge capabilities, thus offering a viable alternative for training LLMs. As a result, our results show promising signs that LLMs using Hyena as a drop-in replacement for attention, coupled with progressive knowledge transfer, are more computationally efficient during model training, compared to current attention-based transformers.

\newpage
\bibliography{writeUp}
\bibliographystyle{icml2024}

\newpage
\appendix
\onecolumn
\section{Appendix}

\subsection{Hyper Parameters}

Hyperparameter tuning played a pivotal role in optimizing the distillation process. Tuning focused on the learning rate and batch size for the generated activations of the teacher model. Three values for each variable were systematically tested, with the selection based on achieving the lowest Mean Squared Error (MSE) for the 6th layer of the distilled model. The resulting validation and training losses are summarized in Table \ref{hyperparameter-search}.

\begin{table}[h]
\caption{Distillation hyper parameter search results}
\label{hyperparameter-search}
\vskip 0.15in
\begin{center}
\begin{small}
\begin{sc}
\begin{tabular}{c c c}
\toprule
\textbf{(Learning rate, batch size)} & \textbf{Training MSE} & \textbf{Validation MSE}  \\
\midrule
$(0.001, 60)$ &  \textbf{0.1312} & \textbf{0.1344}\\
\hline
$(0.0025, 60)$ &  0.1669 & 0.1652\\
\hline
$(0.0001, 60)$ &  0.2050  & 0.2012 \\
\hline
$(0.001, 240)$ & 0.3111 & 0.3069 \\
\bottomrule
\end{tabular}
\end{sc}
\end{small}
\end{center}
\end{table}

\begin{table}[ht]
    \caption{Best hyper-parameters for the 2 methods of distillation}
    \label{pre_trained_hp}
    \vskip 0.15in
    \begin{center}
    \begin{small}
    \begin{sc}
    \begin{tabular}{lcccr}
    \toprule
    Hyper-parameter & MSE & CE fine-tune \\
    \midrule
    Distillation epochs    & 8 & 6\\
    Fine-tuning epochs & 0 & 6\\
    Weight decay    & 0.1 & 0.1\\
    Maximum learning rate    & $1 \cdot 10^{-3}$ & $1 \cdot 10^{-3}$\\
    Minimum learning rate & 1 $\cdot 10^-4$ & 1 $\cdot 10^-4$\\
    Betas     & (0.9,0.98) & (0.9,0.98)\\
    Batch size & 60 & 60\\
\bottomrule
\end{tabular}
\end{sc}
\end{small}
\end{center}
\vskip -0.1in
\end{table}

\begin{table}[ht]
    \caption{Best hyper-parameters for the pre-trained model}
    \label{from_scratch_hp}
    \vskip 0.15in
    \begin{center}
    \begin{small}
    \begin{sc}
    \begin{tabular}{lc}
    \toprule
    Hyper-parameter & Value  \\
    \midrule
    Weight decay    & 0.1 \\
    Maximum learning rate    & $1 \cdot 10^{-3}$ \\
    Minimum learning rate & 1 $\cdot 10^-4$ \\
    Betas     & (0.9,0.98) \\
    Warm-up percentage & 2.5\%\\
    Total number of tokens & 1B\\
    Batch size (tokens) & 0.5M \\
\bottomrule
\end{tabular}
\end{sc}
\end{small}
\end{center}
\vskip -0.1in
\end{table}

\end{document}